\documentclass[pmlr,twocolumn,10pt]{jmlr} 

\usepackage{booktabs}
\usepackage{siunitx}

\usepackage{colortbl}
\usepackage{soul}
\definecolor{lightred}{rgb}{1, 0.9, 0.9} 
\definecolor{lightgreen}{rgb}{0.9, 1, 0.9} 

\usepackage[switch]{lineno}
\usepackage [english]{babel}
\usepackage [autostyle, english = american]{csquotes}
\MakeOuterQuote{"}

\newcommand{\equal}[1]{{\hypersetup{linkcolor=black}\thanks{#1}}}

\theorembodyfont{\upshape}
\theoremheaderfont{\scshape}
\theorempostheader{:}
\theoremsep{\newline}

\jmlryear{2024}
\jmlrworkshop{Conference on Health, Inference, and Learning (CHIL) 2024} 

\title[Regulating AI Adaptation]{Regulating AI Adaptation: An Analysis of AI Medical Device Updates
}

\author{%
\Name{Kevin Wu}\equal{Equal first authorship} \Email{kevinywu@stanford.edu}\\
\addr Stanford University, USA
\AND
\Name{Eric Wu}\footnotemark[1] \Email{wue@stanford.edu}\\
\addr Stanford University, USA
\AND
\Name{Kit Rodolfa} \Email{krodolfa@law.stanford.edu}\\
\addr Stanford University, USA
\AND
\Name{Daniel E. Ho}\equal{Equal senior authorship} \Email{dho@law.stanford.edu}\\
\addr Stanford University, USA
\AND
\Name{James Zou}\footnotemark[2] \Email{jamesz@stanford.edu}\\
\addr Stanford University, USA
}

\begin{document}

\maketitle

\begin{abstract}
While the pace of development of AI has rapidly progressed in recent years, the implementation of safe and effective regulatory frameworks has lagged behind. In particular, the adaptive nature of AI models presents unique challenges to regulators as updating a model can improve its performance but also introduce safety risks. In the US, the Food and Drug Administration (FDA) has been a forerunner in regulating and approving hundreds of AI medical devices. To better understand how AI is updated and its regulatory considerations, we systematically analyze the frequency and nature of updates in FDA-approved AI medical devices. We find that less than 2\% of all devices report having been updated by being re-trained on new data. Meanwhile, nearly a quarter of devices report updates in the form of new functionality and marketing claims. As an illustrative case study, we analyze pneumothorax detection models and find that while model performance can degrade by as much as 0.18 AUC when evaluated on new sites, re-training on site-specific data can mitigate this performance drop, recovering up to 0.23 AUC. However, we also observed significant degradation on the original site after re-training using data from new sites, providing insight from one example that challenges the current one-model-fits-all approach to regulatory approvals. Our analysis provides an in-depth look at the current state of FDA-approved AI device updates and insights for future regulatory policies toward model updating and adaptive AI.
\end{abstract}

\paragraph*{Data and Code Availability}
The primary data used in this study are publicly available through the FDA website. Our analysis of the data and code used is available in the supplementary material and will be made publicly available on GitHub.

\paragraph*{Institutional Review Board (IRB)}
Our research does not require IRB approval.

\begin{figure*}[h]
\centering
\includegraphics[width=0.6\textwidth]{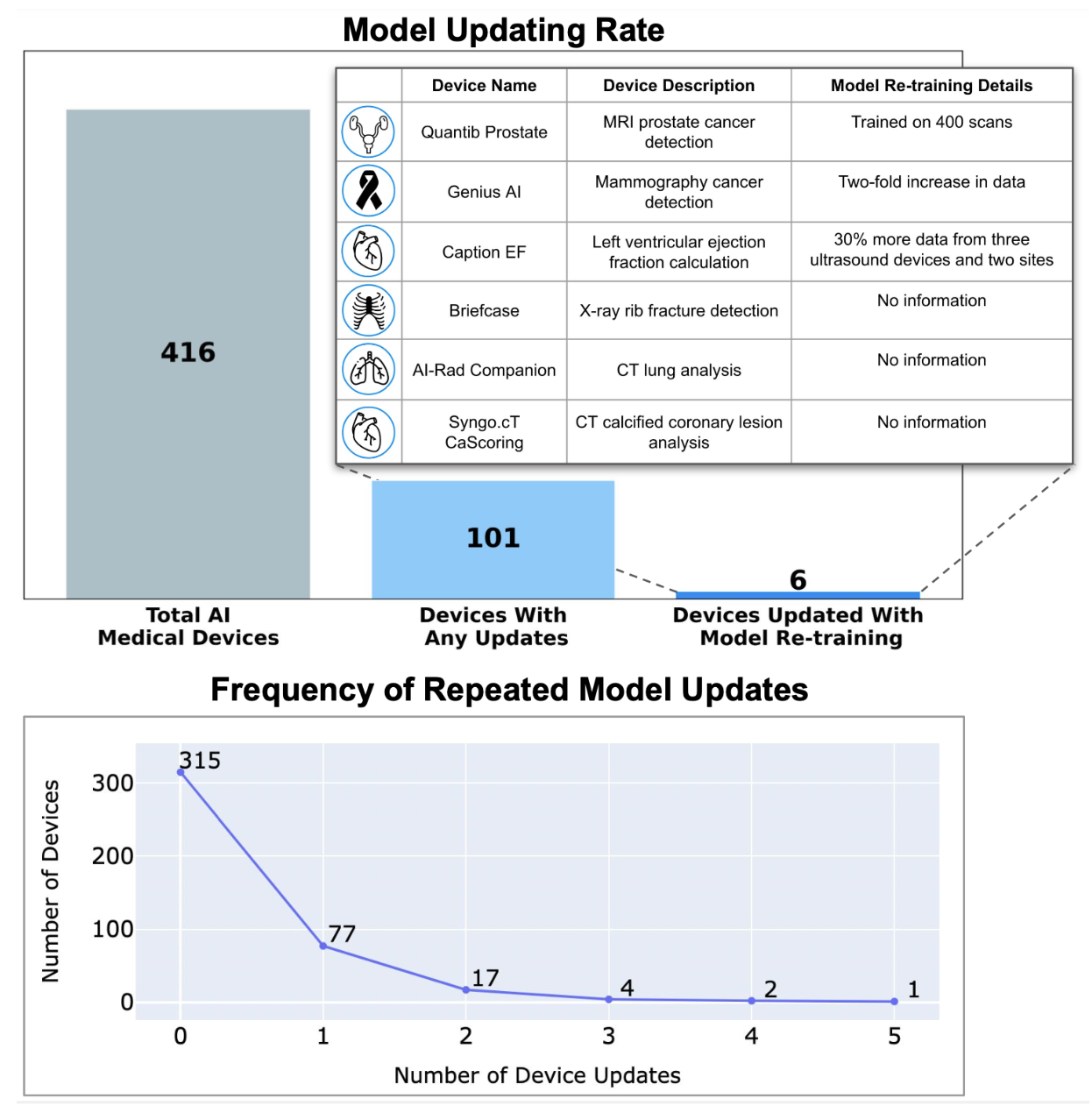}
\caption{(Top) Proportion of devices that report model re-training and other update types. The devices with any updates are a subset of the total AI medical devices, and the devices updated with model re-training are a subset of devices with any updates. The call-out provides details on the six devices that received model re-training, along with device name, device description, and details provided within the FDA approval regarding the type of model re-training applied. (Bottom) Graph of the number of times devices have been updated. The x-axis refers to the number of successive updates, and the y-axis refers to the count of devices in each group.}
\label{fig:figure1}
\end{figure*}

\section{Introduction}
\label{sec:intro}
While the number of AI products developed for commercial applications is rapidly growing, the implementation of robust regulatory frameworks still lags behind \citep{Larson2021-ng,Wirtz2020-ww,Wu2021-ui}. Recently, high-profile accidents involving Boeing \citep{Wendel2019-vj} and Tesla \citep{Corfield2023-gp} have been attributed to issues with software and AI updates in their systems. Applications of AI to consumer lending \citep{Johnson2019-dj} and hiring systems \citep{Bogen2018-fx} has also led to calls for more flexible regulatory systems that can anticipate algorithmic changes and biases. Such cases highlight the inherent challenges regulators face due to the adaptive nature of software and especially AI products: while model adaptation and updates are a necessary step in maintaining or improving their performance, they can also introduce unknown safety risks \citep{Babic2019-qe,Gilbert2021-bh}.

\begin{figure*}[ht]
\centering
\includegraphics[width=0.7\textwidth]{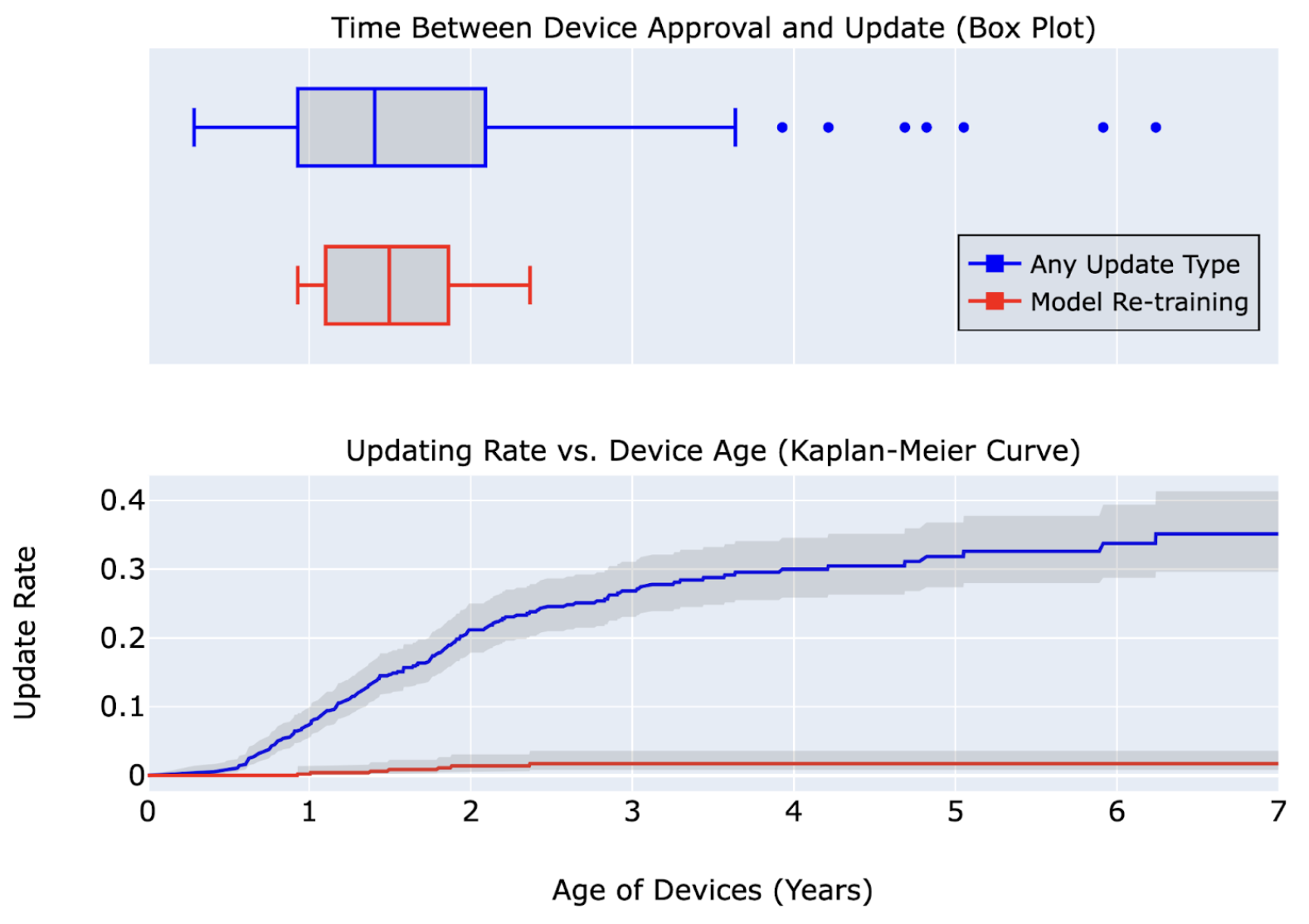}
\caption{(Top) Time to device update (in years), represented as a box plot where each vertical bar represents the 25th and 75th percentile, respectively. The red line represents the distribution of time to update for model re-training, while the blue line refers to time to update of any type. (Bottom) The estimated rate of updating as a function of device age, which is estimated through the Kaplan-Meier function in order to account for the right censorship in the dataset. For example, devices at two years old are updated across all types approximately 20\% of the time, whereas the same-age devices are updated with model re-training around 1.4\% of the time. Both plots have been cut-off at seven years as the longest device update observed is 5.9 years. Shading indicates the 95\% confidence interval around the updating rate at any given time. }
\label{fig:figure2}
\end{figure*}

In the US, the Food and Drug Administration (FDA) has been an early mover in AI regulation, with over 500 approved submissions for AI devices as of 2022 \citep{Center_for_Devices2022-qx}. The FDA faces unique challenges with regard to model updating, as adverse events can directly compromise patient well-being. As such, the FDA has traditionally not allowed any changes to a model once it has been approved \citep{Gerke2020-bf}. At the same time, AI models are well-known to be prone to distribution shifts, whereby variations in factors such as medical practice, patient demographics, or disease prevalence can significantly affect a model's performance \citep{Raghu2019-dr,Wiens2019-qt,Wong2021-rs}. For example, researchers recently found that Epic's widely used sepsis prediction model performed much worse than initially reported after being deployed in a new hospital setting \citep{Wong2021-rs}. Such cases demonstrate that fixed AI models that never receive updates can likewise compromise patient safety. Recently, the FDA has taken action to address the limitations of a fixed-model approach by providing guidelines for a potential Predetermined Change Control Plan (PCCP) \citep{Center_for_Devices2023-av}, as well as a document describing best practices in machine learning published jointly by US, Canadian, and UK health authorities \citep{Center_for_Devices2023-vu}. Under this provision, developers can make a limited set of changes to their models without a new submission as long as it is pre-specified in their initial approval. Such proposed measures by the FDA underscore the importance of the ongoing discussion around the appropriate levels of regulation regarding the adaptive nature of AI.

Despite the importance of model updating in AI medical devices, little is known about how often such devices are currently being updated. While AI developers may individually publish press releases about changes to their model, there does not exist a systematic analysis of updating across all AI medical devices. FDA approvals by the same developer often contain variants of company and product names, making it difficult to automatically link devices together. Furthermore, devices under the same name often vary widely according to their use cases and are actually different products. In this study, we aim to resolve these issues by organizing and grouping FDA-approved AI medical devices by their updates and performing an analysis of the frequency and nature of model updates. Our study explores the extent to which developers choose to update their devices given current regulatory, economic, and technological factors. Additionally, we perform an illustrative case study on AI models designed to predict pneumothorax, evaluating whether model updates consistently yield improved performance when re-trained on target populations.

\section{Methods}

\subsection{Collecting device updates}

The primary data for this study consists of FDA approval documents for AI medical devices, which are publicly available through the FDA's online database (\href{www.fda.gov}{www.fda.gov}). Under the FDA's 510(k) approval process, developers must demonstrate that the medical device they are marketing (the "subject" device) is "substantially equivalent" to a device already available on the market (the "predicate" device) \citep{Brindza1980-xm}. Furthermore, each FDA-approved device is classified using a product code that indicates the overall function and safety profile \citep{Center_for_Devices2023-av}. For example, the product code QFM refers to "Radiological Computer-Assisted Prioritization Software For Lesions" and includes many common triage-based AI detection software. In our analysis, an FDA approval is considered a device update if 1) the predicate and the subject devices are from the same manufacturer, 2) both devices share the same product classification code, and 3) both devices are AI devices.

When grouping by manufacturer names, multiple variants of the same manufacturer often appear (e.g., Siemens Medical Solutions USA Inc. and Siemens Medical Solutions, Inc.). To reconcile these differences, we first applied approximate string matching with Levenshtein distance and a similarity threshold of 0.8 to create candidate company name groupings before manual review. Furthermore, to systematically identify the predicate devices for each FDA approval, we extracted the PDF texts and performed a search over the first appearance of a submission number outside of the subject device number before performing a manual review. 

\begin{figure}[h]
\centering
\includegraphics[width=0.4\textwidth]{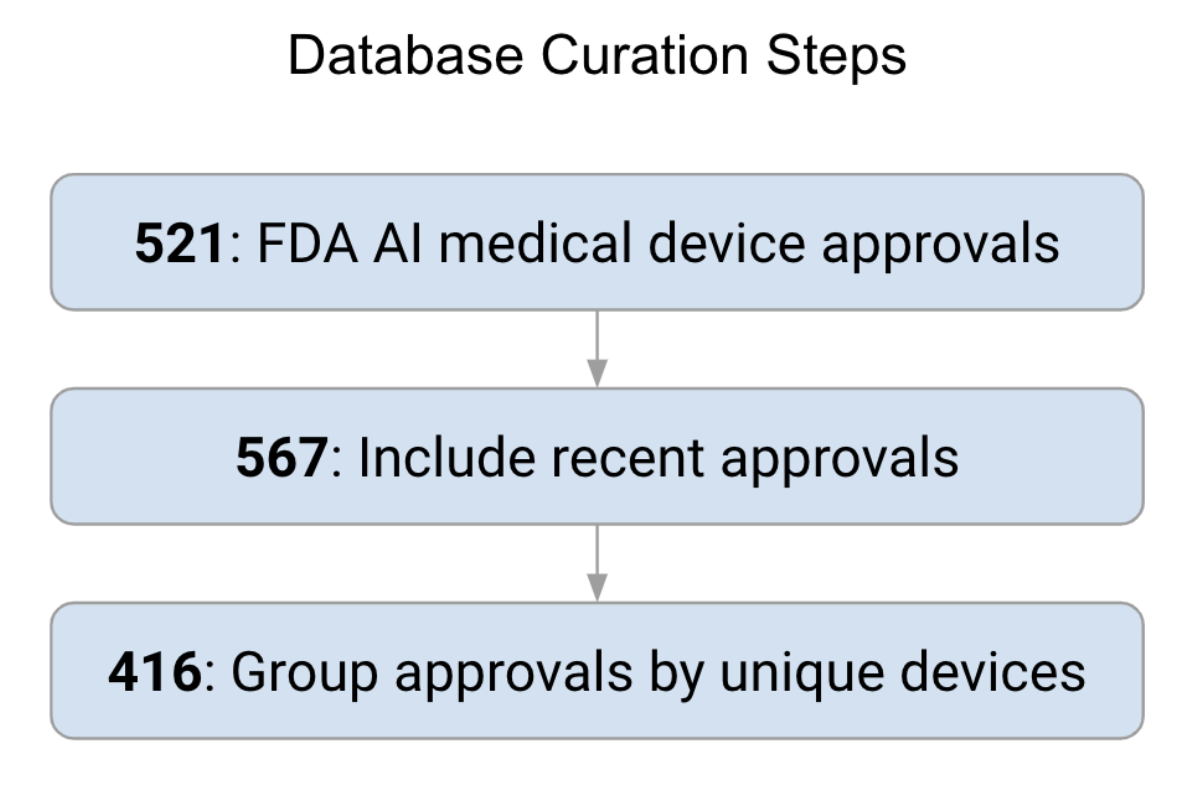}
\caption{Schematic of database curation steps, starting from the FDA's official list of AI medical device approvals to the final set of 416 unique devices.}
\label{fig:suppfigure1}
\end{figure}

\begin{table*}[ht]
\centering
\begin{tabular}{p{3.7cm}p{3cm}p{3.8cm}p{3.8cm}}
\toprule
\textbf{Update Example} & \textbf{Example FDA Approval} & \textbf{Example Product} & \textbf{Example Change} \\ \hline
Model design & K201310 & Caption Health & Simplification of network architecture via simple pooling layers and quantization \\ \midrule
Input signal & K211541 & Mammoscreen & Added 3D (DBT) mammogram support \\ \hline
Target population & K210034 & EnsoSleep & Pediatric Patient Population Included \\ \hline
Training data & K221240 & BriefCase & Larger training dataset \\ \hline
Device User & K203508 & BriefCase & From radiologist only to medical professionals \\ \hline
Compatibility & K191994 & ProFound AI Software V2.1 & Added support for Siemens machines \\ \hline
UI/UX & K221240 & BriefCase for IHC Triage & New desktop application \\ \hline
Hardware & K221147 & Vivid T8, Vivid T9 & Modified transducers \\ \hline
Software & K220590 & aPROMISE X & Accessible through cloud \\ \hline
Device Use Case & K193417 & FractureDetect (FX) & Expanded range of fracture detection from wrist to ankle, elbow, etc. \\ \bottomrule
\end{tabular}
\caption{Examples of updating types present in follow-up devices. The table provides the update type, along with an example of each subtype.}
\label{table:table1}
\end{table*}

Our dataset starts with the FDA's list of AI/ML medical devices, which contains a total of 521 FDA approvals (recent as of 10/5/22). In order to include more recent updates, we added an additional 46 approvals from 10/5/22-07/01/23 that reference one of the 521 approvals as a predicate device. In total, our final dataset contains 416 unique devices, which are represented across 567 total FDA approvals (e.g. a single device can be approved multiple times for each update). The data curation steps and sample sizes are outlined in Figure \ref{fig:suppfigure1}.

After identifying all device updates, we determined the types of updates that occur. For each FDA approval, manufacturers are required to provide details of the subject device's technological comparison to the predicate device. For example, FDA approval K221727 (syngo.CT Extended Functionality) includes a section titled "Comparison of Technological Characteristics with the Predicate Device", which contains a table comparing and contrasting the predicate (SOMARIS/8 VB60) with the subject device (SOMARIS/8 VB70). Within this section, the update is described to have "Improved quality of the bone removal algorithm for the head \& neck region", and notes that "Segmentation of the bones use a deep learning algorithm instead of a traditional image processing". We annotated each updated device according to the type of update received, which is further detailed in the Results section.

\begin{figure*}[ht]
\centering
\includegraphics[width=0.7\textwidth]{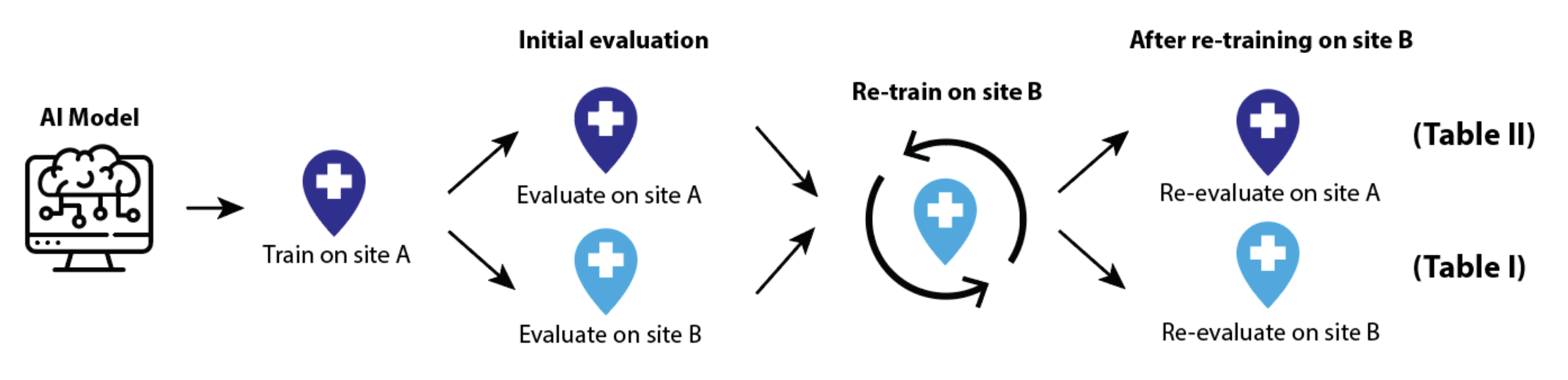}

\bigskip 

\begin{minipage}{\textwidth}
\centering
\begin{tabular}{|l|l|l|l|}
\hline
\multicolumn{1}{|c|}{\textbf{Train on site A}} & \multicolumn{3}{c|}{\textbf{Evaluate on site B (before and after re-training)}} \\ \hline
 & \textbf{SHC (N=39642)} & \textbf{BID (N=71238)} & \textbf{NIH (N=17417)} \\ \hline
SHC & 0.851 & \cellcolor{lightgreen}0.830 → 0.867*** & \cellcolor{lightgreen}0.823 → 0.835 \\ \hline
BID & \cellcolor{lightgreen}0.799 → 0.821*** & 0.879 & \cellcolor{lightgreen}0.782 → 0.850*** \\ \hline
NIH & \cellcolor{lightgreen}0.727 → 0.809*** & \cellcolor{lightgreen}0.645 → 0.883*** & 0.848 \\ \hline
\end{tabular}
\end{minipage}

\bigskip 

\begin{minipage}{\textwidth}
\centering
\begin{tabular}{|l|l|l|l|}
\hline
\multicolumn{1}{|c|}{\textbf{Train on site A}} & \multicolumn{3}{c|}{\textbf{Evaluate on site A (before and after re-training on site B)}} \\ \hline
 & \textbf{SHC (N=39642)} & \textbf{BID (N=71238)} & \textbf{NIH (N=17417)} \\ \hline
SHC & 0.853 & \cellcolor{lightred}0.853 → 0.688*** & \cellcolor{lightred}0.853 → 0.632*** \\ \hline
BID & \cellcolor{lightred}0.878 → 0.717*** & 0.878 & \cellcolor{lightred}0.878 → 0.610*** \\ \hline
NIH & \cellcolor{lightred}0.835 → 0.739*** & \cellcolor{lightred}0.835 → 0.691*** & 0.835 \\ \hline
\end{tabular}
\end{minipage}

\caption{A case study on the effect of re-training pneumothorax AI models on other sites reveals that although re-training improves external site performance, performance consequentially degrades on the originally trained sites. \textbf{Top Figure:} An AI model is trained on pneumothorax cases from site A and then evaluated on held-out cases from site A and an external site (site B). Then, the model is updated by fine-tuning on 5K additional cases from site B and re-evaluated on sites A and B. This procedure is performed for three clinical sites (SHC, BID, NIH) across six total scenarios. The results of re-training and re-evaluation are shown in B and C. \textbf{Middle Table:} each cell shows the AUROC scores of the model evaluated on site B before and after re-training on site B. On average, models improved by 0.075 AUC after re-training. \textbf{Bottom Table:} each cell shows the AUROC scores of the model evaluated on site A before and after re-training on site B. Across both panels, we perform bootstrapped one-sided tests for each cell and indicate with asterisks (***) where $p<0.001$.
}
\label{fig:figure3}
\end{figure*}

\subsection{Case study}

Given that site-specific re-training is not allowed under current FDA 510(k) guidelines, we conducted a case study on pneumothorax detection models for chest X-rays to understand the potential performance gains that are currently uncaptured. There are currently four FDA-approved medical devices for the triage of X-ray images for the presence of pneumothorax \citep{Wu2021-ui}, and there are multiple publicly available chest X-ray datasets that include pneumothorax as a condition. We used three datasets, each from a different hospital site in the USA: the National Institutes of Health Clinical Center in Bethesda, Maryland (NIH) \citep{Wang2017-hg}; Stanford Health Care in Palo Alto, California (SHC) \citep{Irvin2019-ji}; and Beth Israel Deaconess Medical Center in Boston, Massachusetts (BID) \citep{Johnson2023-ft}. We used a DenseNet-121 deep-learning architecture \citep{Huang2017-bq} that has been demonstrated to be a top-performing model for the classification of chest conditions \citep{Irvin2019-ji,Seyyed-Kalantari2020-yv}. These datasets represent a diversity of patient populations, imaging manufacturers, and pathology reporting standards \citep{Wu2021-mp}. To quantify how the AI’s performance varies across sites, we trained separate deep-learning models on data from patients at each of the three sites and then evaluated the models on the test set from the other two sites. Each model takes as input a chest X-ray image and makes a binary prediction for pneumothorax. Similar to top-performing model approaches \citep{Irvin2019-ji,Seyyed-Kalantari2020-yv}, we trained five identical models (with different random seeds) for each setting and then ensembled the predictions by averaging the predicted probabilities across each model. We then re-trained the model (by fine-tuning) on a small subset of training data of five thousand examples from an unseen external site and re-evaluated the model's performance on both the original and external sites. We perform fine-tuning with the standard approach of updating all the model weights for a fixed number of steps without changing the hyperparameters.

\section{Results}
\label{sec:results}

\subsection{Device Update Frequency and Types}

Among our dataset of 416 unique devices, we found that 101 devices report having been updated at least once (Figure \ref{fig:figure1}). However, the vast majority of these updates expand the functionality or marketing claims of the device, essentially constituting a new device rather than a true model update. Of these 101 devices, only six of the updated devices report retraining in the model with new data. For each of the six devices, details on the types of data used in re-training are limited, with only three providing how much training data was used. For AI devices, retraining on new data is central to and distinctive of the technology, leading to our focus on the novel regulatory issues here. For example, Syngo.CT CaScoring (K221219), which analyzes calcified coronary lesions, only references that "the algorithm was re-trained on a larger database". AI-Rad Companion (K213096), which analyzes lung CTs, references "additional training data was added", while Briefcase (K230020), a rib fracture triage device, mentions that the updated device differs "due to training the subject device on a larger data set". The remaining three devices reference the scale of the re-training dataset. For example, Quantib Prostate (K230772), which analyzes prostate MRIs, reports that the updated algorithm has been trained on "400 scans", while Genius AI (K221449), a breast cancer detection device, reports a "two-fold" increase. Finally, Caption Ejection Fraction (K210747), a cardiac ultrasound AI device, reports an "additional 30\% training data from three ultrasound devices and two clinical sites". Details on these devices are also included in Figure \ref{fig:figure2}. 
For the other 95 updated devices, we found several different update subtypes. The most common type of reported updating occurs when the manufacturer adds a new or additional prediction task to an existing model (55 total devices). For example, whereas FractureDetect's original device only works on wrists, its update has expanded to ankles, elbows, and other body parts. Next, we found that 21 devices have received updates to their accepted input signal. For example, recent mammography products such as Mammoscreen have included the ability to process Digital Breast Tomosynthesis (DBT)/3D scans, whereas previous versions only accepted Full-Field Digital Mammography (FFDM)/2D scans. An additional 13 devices report changes to the model design or architecture, such as a change from a fully connected neural network to a convolutional neural network. Five devices report a change to the intended target population for the device. For example, EndoSleep expanded its population to pediatric patients, whereas the previous device only allowed for patients 18 or older. We found 22 devices that report changes to the model but do not specify the exact nature of the change. For example, approvals may report "additional algorithmic enhancements" or "improved quality of algorithms", but not reference whether the improvements come from re-training or model design. Finally, 37 devices report updates unrelated to the model or its usage. Namely, these include software or hardware changes that pertain to its interoperability or output interface. Examples include the UI/UX of the device which is visible to physicians, or a hardware configuration that allows the device to be installed on new machines. We provide a list of examples of these update types in Table \ref{table:table1}.

\subsection{Time Between Updates}

Based on our dataset, updates of any type occur a median of 17 months after previous device approval, with follow-ups as short as 3.5 months and as long as six years (Figure \ref{fig:figure2}). This is a relatively short window of time, as the median time from concept to FDA approval for non-AI medical devices has been estimated to be 31 months (C. Johnson et al., 2022). Additionally, in order to account for right-censorship in our dataset (e.g. not yet observed updates in the newer devices), we used the Kaplan-Meier estimator and produced its curve (Figure \ref{fig:figure2}). At two years, devices have an estimated update probability of 20\% for all update types, and at four years, this probability rises to 30\%. After seven years, the estimated probability of update saturates at 35\%, meaning that about a third of devices receive at least one update of any kind in their lifetimes. However, the reported rate of model re-training is significantly lower: within two years, 1.4\% of models are reported to be retrained, with the probability of device updates saturating at 1.7\% after 2.4 years. 

\subsection{Case Study}

We carried out a case study to illustrate and quantify the tradeoffs with AI adaptation through model retraining. We investigated the potential benefits and challenges of re-training on additional data from external sites in pneumothorax AI algorithms \citep{Wu2021-mp, Wu2021-ui}. We found that external evaluation of models can result in an AUC decrease of up to 0.18, while re-training and evaluating on data from external sites improves model performance in all scenarios, with an average of 0.075 and a maximum of 0.23 AUC (Figure \ref{fig:figure3}, Middle). However, after re-training on external sites, we also found that model performance degrades an average of 0.176 AUC (and up to 0.268 AUC) when re-evaluated on the original site (Figure \ref{fig:figure3}, Bottom). This suggests that it can be challenging to have a single AI model that works well across heterogeneous settings.

\section{Discussion}

Currently, FDA-approved AI models are "locked" after approval, whereby making new changes requires undergoing a brand-new submission process, with most of the same regulatory burden \citep{Gerke2020-bf}. Correspondingly, we observe in our analysis that only six out of 416 devices report actually received re-training updates, which is an essential approach for AI adaptation. On the other hand, nearly a quarter of devices receive updates in the form of additional marketing or functionality claims. Such disparity suggests a much stronger economic incentive for developers to increase the adoption of their devices through marketing new features rather than improving the original model through re-training. 
\\ \\
One significant barrier to re-training is development costs, which may include acquiring new datasets \citep{Chen2019-tk, Wu2023-yb}, computational resources \citep{Wiens2019-qt}, data groundtruthing \citep{Rahimi2021-aj,Willemink2020-mc}, and regulatory hurdles \citep{Kelly2019-hu, Sertkaya2022-be}. After models are updated, the manner in which they are deployed can also affect a device's ultimate clinical impact. First, while previous-generation AI devices for mammography were clinically evaluated to improve detection rates, subsequent studies showed limited benefits to women due to changes in how clinicians interacted with the devices, as well as the transition from film to digital mammograms \citep{lehman2015diagnostic, fenton2015time}. Second, economic forces such as reimbursement rates can affect how the frequency and extent to which these devices are adopted \citep{parikh2022paying, abramoff2022reimbursement}. AI adoption is still in a nascent stage, with very few widely adopted products and underdeveloped commercial payment pathways \citep{Chen2021-ww, Parikh2022-rp, Wu2023-ri}. In such an environment, companies with few customers may not be able to dedicate resources toward regular model updating and maintenance. Currently, FDA cleared products exist in a similar band of risk profiles, with a previous study showing all devices currently categorized as risk class II (medium-risk) (\citep{zhu20222021}). The lower-risk class I is largely exempt from the regulatory process, with the higher-risk class III reserved for devices that "sustain or support life, are implanted, or present potential unreasonable risk of illness or injury" (\citep{fda_guidance}). Whereas minimal-risk products like mobile health apps can introduce frequent updates without any regulatory hurdles, the medium-risk designation may encourage a trend towards more conservative updates that are more likely to be cleared rather than ambitious updates that may be rejected. 
\\ \\
The FDA has recognized the high regulatory hurdles associated with model updating. In a recently proposed draft guidance from April 2023, model developers may be allowed to include a PCCP (Predetermined Change Control Plan) along with their device submission, which would allow them to simply document subsequent model updates rather than requiring a new submission every time \citep{Center_for_Devices2023-ug}, potentially alleviating some of the regulatory burden and shortening update intervals. However, even under these proposed changes, developers are still required to complete rigorous evaluation and documentation of the algorithm changes, which incur much of the same prohibitive time and costs mentioned above \citep{Allen2022-my, Evans2022-rz}. Furthermore, evaluating an updated model is an inherently difficult task due to various types of distribution shifts and heterogeneous data collection methods that are outside the control of developers \citep{Schrouff2022-ma, chen2018my}. As such, future guidance documents should consider the challenges inherent in ensuring and evaluating fairness under fine-tuning and data shift.
Our case study illustrates the tug-and-pull nature observed in AI models -- when trained on data from a specific site, they can perform well, but this may trade-off with performance on other sites. Although our models are trained on only a few datasets and do not comprehensively represent the gamut of available training data sources and model architectures on the market, the results illustrate how one instantiation with commonly used data and architecture choices exhibits characteristic behaviors of performance shift. In the status quo, model developers are locked into one model, creating scenarios where they may have to optimize for one population at the expense of another. To compound this issue, the actual performance on new, unseen populations is not even reported since the FDA does not require postmarket surveillance for 510(k)-approved devices \citep{Wu2021-ui}.
To alleviate this problem, future regulatory guidelines should move beyond a "one-model-fits-all" approach, and instead consider allowing site-specific re-training and deployment. By allowing developers to deploy and validate multiple models under a single device, they can optimize model performance for each intended population without incurring performance tradeoffs. This would ensure that developers verify that their models perform well on each deployed clinical site while allowing them to perform the necessary site-specific documentation and evaluation as they mature.
There are various design decisions that can affect how an AI model is re-trained: factors like whether to freeze layers, mix new training data, hyperparameter tuning, and validation processes can all influence how much re-training improves model performance \citep{Pham2021-wf, Picard2021-nv, Qian2021-zd}. Indeed, in our case study, even though the individual models used in our ensemble approach only varied by the random seed used during training, performance across models still differed by up to 0.056 AUC. In a study by \citet{Watson2022-co}, chest X-ray deep learning models trained on the same BIDMC dataset across different random seeds and hyperparameters were found to disagree in their explanations up to two-thirds of the time. Such studies on specific datasets represent potential pitfalls of algorithms applied to a particular clinical domain, but do not necessarily mean they generalize to all other types of devices. Regulatory guidelines should include consideration of appropriate fine-tuning schemes used when evaluating models.

Furthermore, we find that among models that have been updated with re-training, details on the data used in training are very limited, with basic descriptions such as "additional training data", or "larger database". A limitation of our study lies in the limited details reported in FDA clearances. For example, while only 6 devices report retraining on new data, 5 devices report updates to their target population and 21 devices report unspecified improvements to their algorithm. As such, the true rate of retraining on new data may be higher than reported. In order for consumers and users to make informed decisions on the impacts of model updates, regulators should ensure that information about the data used for original model development and updates is transparent and accessible. Information such as patient demographics, hospital locations, disease subtypes, and healthcare settings are important covariates that can significantly influence model performance \citep{Duffy2022-iu, Wu2021-mp}.

As the medical AI field matures, regulation should progress in lockstep with fully exploiting the technical benefits of adaptive learning systems while curbing risks to safety and efficacy. Looking beyond the US, regulatory bodies, such as the European Union, are similarly developing guidelines for regulating medical AI \citep{Muehlematter2021-mi}. We believe that the trends and challenges in medical AI also extend to regulating other AI-transformed industries such as transportation and law, and offer important insights into how to appropriately foster AI innovation.

\bibliography{chil-sample}

\end{document}